\documentclass[doublecol]{epl2} 
\usepackage{kotex}

\title{Very simple statistical evidence that AlphaGo has exceeded human limits in playing GO game}
\shorttitle{Title} 

\author{Okyu Kwon\inst{1}}

\institute{                    
  \inst{1} Division of Medical Mathematics, National Institute for Mathematical Sciences, Daejeon 34047, Korea \\
}

\abstract{
Deep learning technology is making great progress in solving the challenging problems of artificial intelligence, hence machine learning based on artificial neural networks is in the spotlight again. In some areas, artificial intelligence based on deep learning is beyond human capabilities. It seemed extremely difficult for a machine to beat a human in a Go game, but AlphaGo has shown to beat a professional player in the game. By looking at the statistical distribution of the distance in which the Go stones are laid in succession, we find a clear trace that Alphago has surpassed human abilities. The AlphaGo than professional players and professional players than ordinary players shows the laying of stones in the distance becomes more frequent. In addition, AlphaGo shows a much more pronounced difference than that of ordinary players and professional players.
}

\begin{document}

\maketitle

\section{Introduction}
Deep learning is a class of machine learning based on artificial neural networks. In the case of artificial neural networks, stacking four or more layers has been treated as meaningless. However, a method has proposed to effectively train artificial neural networks stacked with a large number of layers, increasing the computational power of computers with powerful GPUs, and a large amount of digital data to be used for learning has brought back the revival of machine learning based on artificial neural networks since 2000. Deep learning is making great achievements in solving challenging problems in the field of artificial intelligence (AI)\cite{lecun2015}. It has outperformed other machine learning technologies in various areas, such as identifying objects in images, transcribing speech into text, recognizing human speech, drug design, medical image analysis, material inspection, autonomous driving, and playing board game. There are even situations that surpassing human experts in some cases\cite{ciregan2012, krizhevsky202, gibney2016}.

The chess skills of the computing machine have already surpassed humans.  In 1997, IBM's supercomputer Deep Blue beat the chess world champion, Garry Kasparov\cite{Deepblue}. The ability quickly to calculate the number of nearly all possible cases in which a game of chess will be played and the vast amount of memory that can store them were the key to how computers exceeded human chess skills. Go is a hugely complex ancient strategy board game. Go has an unimaginable number of deployment possibilities that cannot be compared with chess. Therefore, according to the way machines won in chess, it was expected that it would take a long time for the machine to overcome the human Go skills.
However, with the remarkable development of deep learning technology, the AI Go program called AlphaGo was completed\cite{AlphaGo, Silver2016}, and in March 2016, the battle between AlphaGo and the world's best professional Go player Lee Sedol was broadcast all over the world\cite{gibney2016}. AlphaGo surprised the world by winning 4-1 against Lee Sedol. It was a historical event that prove to people around the world that the computing machine exceeded human ability in Go game. It's clear that AlphaGo's Go skills based on deep learning are better than humans, but it's hard to tell how much of that level is. Go experts who watched the game of AlphaGo and Lee Sedol and Lee Sedol, who was AlphaGo's opponent, felt the level qualitatively. We have found from the Go game record a very simple statistical sign that AlphaGo's Go skills differ from humans. We found the statistical signs that can distinguish the ordinary people, professional Go player, and AlphaGo, and in particular, the difference AlphaGo versus professional player is much more distinct than the professional player versus ordinary people.

\section{Data}

\begin{table}[h!]
\begin{center}
 \caption{The number of Go game records of TYGEM and TOM by year.}
 \label{table1}
 \begin{tabular}{c|r|r}      
     \textbf{Year} & \textbf{TYGEM} & \textbf{TOM} \\
     \hline
     2003 & - & 237 \\
     2004 & - &  3,629  \\
     2005 &  &  5,098  \\
     2006 & 77,076  &  4,368 \\
     2007 & 82,830  & 5,497 \\
     2008 & 91,042  & 7,547 \\
     2009  & 107,176 &7,502 \\
     2010 & 120,667 & 10,451 \\
     2011 & 142,671  & 6,627 \\
     2012 & 173,405 & -\\
     2013 & 189,840 & - \\
     2014 & 191,115 & - \\
     2015 & 164,618 & - \\
     2016 & 167,085 & - \\
     \hline
    total & 1,516,031 & 50,956
\end{tabular}
\end{center}
\end{table}

\begin{table}[h!]
\begin{center}
 \caption{The number of Go game records for professional players.}
 \label{table2}
 \begin{tabular}{c|r}      
     \textbf{Year} & \textbf{Professional}  \\
     \hline
     1940s & 781 \\
     1950s & 1,442 \\
     1960s &  2,642 \\
     1970s &  3,689 \\
     1980s & 8,055 \\
     1990s & 14,292 \\
     2000s & 21,768 \\
     2010s & 20,853 \\
     \hline
     total & 146,263
\end{tabular}
\end{center}
\end{table}

We obtained Go game record data from GitHub\cite{github}. Go game record data there can be divided into four categories.  One is the record of matches played on the online Go game platform called TYGEM. TYGEM is developed in Korea and is an online platform where anyone who can play Go can participate and enjoy the game of Go. The service started in 2002. The second is the online Go game platform operated in China called TOM. There is a record of games played here.  Third is the record of professional Go players. Professional Go players are people who exceed the ability in Go game of the average people through long hours of training. There is a record of matches between these professional Go players. Fourth is the Go games where AI participated. AlphaGo is not the only AI Go program.  There are various Go algorithms and there is a record of the competition they played. There is also a game record where the algorithm is played against humans. Among them are five games for AlphaGo and Fan Hui and five games for Alphago and Lee Sedol. Go game record data is faithfully accumulated in GitHub. TYGEM data have accumulated 1,516,031 games for the period from November 2, 2005 to December 31, 2016. TOM data have accumulated 50,956 games for the period  from September 25, 2003 to December 28, 2011. Professional data have accumulated 73,522 games from January 1, 1940 to January 9, 2017. The number of games accumulated by year for TYGEM and TOM is shown in Table \ref{table1}. Table \ref{table2} shows the number of games accumulated every 10 years for the professional players.

\begin{figure}
\onefigure[width=0.4\textwidth]{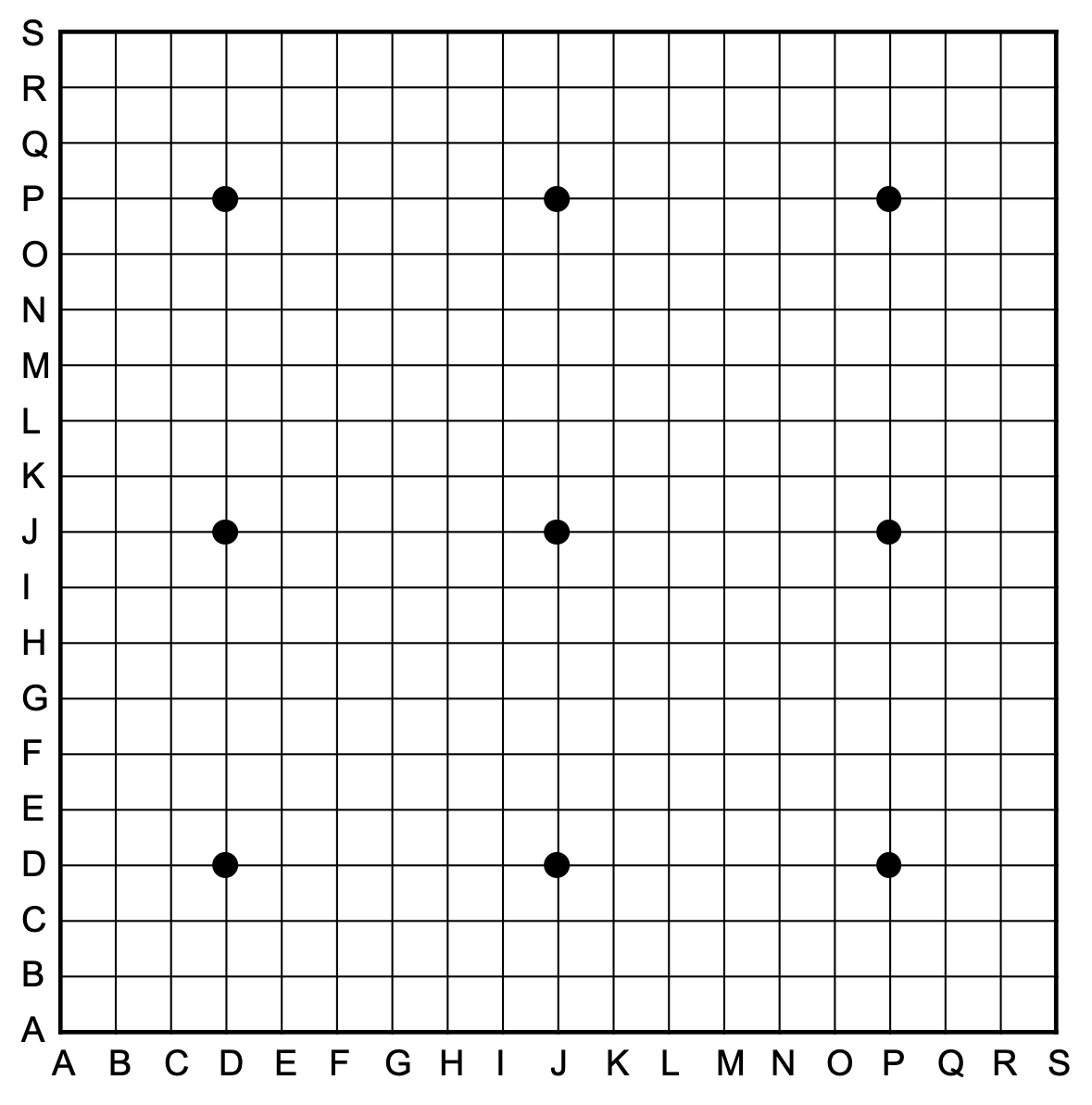}
\caption{Go checkerboard and its coordinates.}
\label{fig1}
\end{figure}

Go is a game where two players compete on a grid of 19 horizontal lines and 19 vertical lines. You can see the Go checkerboard in Fig. \ref{fig1}. One player places a black stone at an empty grid point and the other player places a white stone at an empty grid point. The black and white stones are placed alternately and the game begins with laying black stones. Horizontal and vertical grid coordinates where Go stones are placed are specified by the 19 alphabets from A to S. All grid points on the Go checkerboard can be specified with 361(19x19) coordinates from AA to SS.  The coordinates of grid points in order without missing from the moment the first stone is placed to the moment the last stone is placed is written in Go game record.

\section{Results}
\begin{figure}
\onefigure[width=0.45\textwidth]{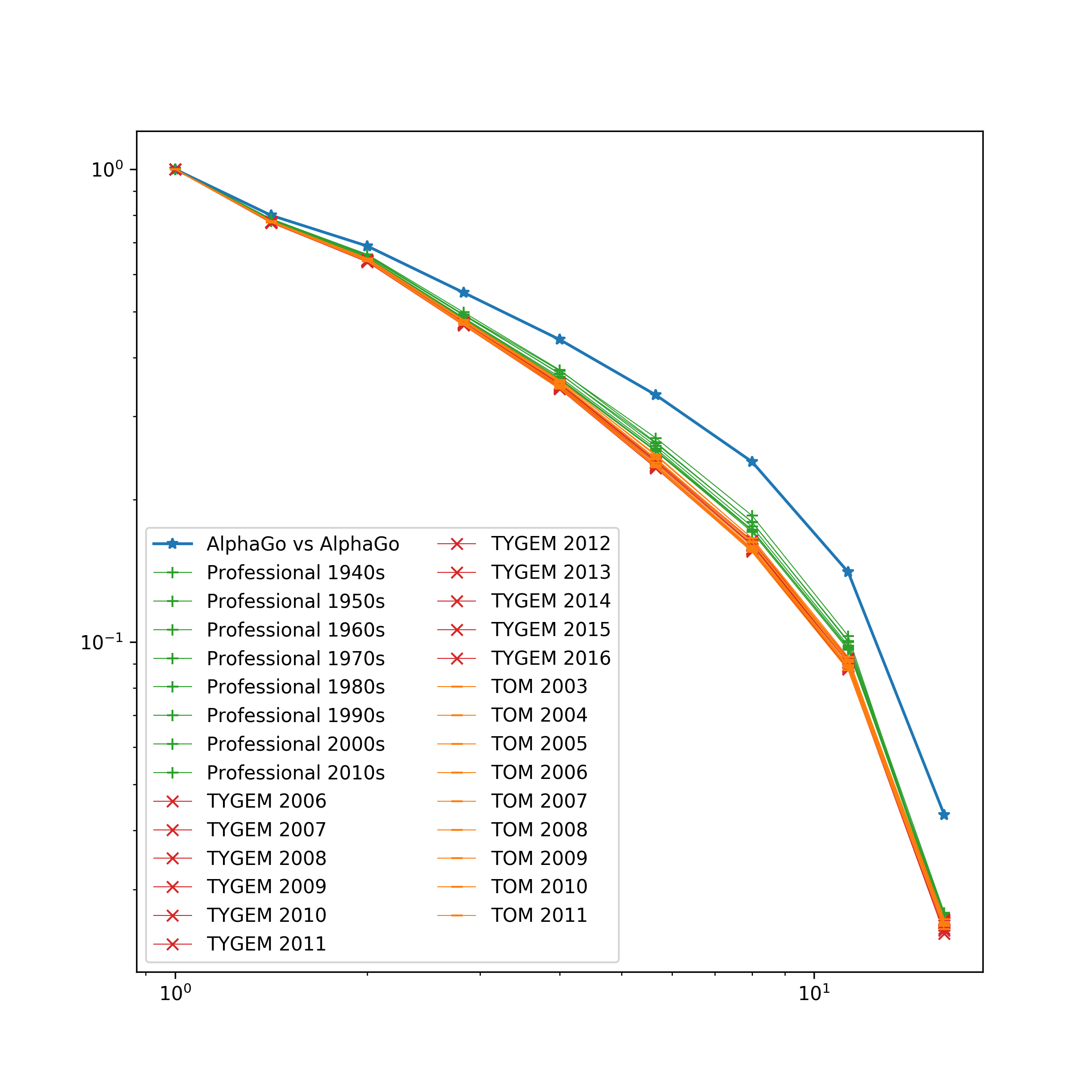}
\caption{The cumulative probability distribution of distances in which black and white stones are placed consecutively over the entire game records separated by period.}
\label{fig2}
\end{figure}

The Go checkerboard is in Cartesian coordinates. Thus, the Euclidean distance between the grid points can be calculated. We can calculate the Euclidean distance between stones placed consecutively from the Go game records. From this, the probability distribution over distances between successive placing stones can be obtained. The results can be seen in Fig. \ref{fig2}. We got a probability distribution from 55 games played by AlphaGo and AlphaGo. The professional competition was divided into every decade periods to obtain a probability distribution over each period. TYGEM's competition was divided into every year to get a probability distribution by year.  TOM's games were also divided by year to obtain their probability distribution. There is a consistent pattern for every Go games in which the probability that the next stone is placed in the near distance is large and the probability that it is placed in the long-distance is reduced. An interesting feature is that the games played in TYGEM and TOM are roughly one group, and the professional player's games are divided into another group in probability distributions. Although the TYGEM's consecutive placement distance probability distribution for each year from 2006 to 2016 and the TOM's such distribution from 2003 to 2011 differs slightly from each other, all patterns form a group. The probability distributions of professional competitions form another group with almost the same pattern, although there is a slight difference from 1940s to 2010s. While there is no dramatic difference, it seems the long-distance placement is happening more frequently in the games of professional Go players than the games played on the online Go platform. The probability distribution from the 55 games of AlphaGo versus AlphaGo clearly shows distinction from the other two distribution groups. You can see that AlphaGo makes long-distance placement much more frequently than those of professional players. It seems that more and more frequently long-range consecutive placement of stone occurs as the game of the average people, professional players, and AlphaGo. Moreover, the difference in the frequency is more noticeable between AlphaGo and human than professional and ordinary people.

In the probability distribution of TYGEM and TOM games, we can see that the tail thickness is almost maintained for almost 10 years. It is also seen in the game of professional players that the thickness of the tail is almost maintained for half a century.  It is believed that the human's Go level has hardly changed during the last half-century or decade. Although the difference between the level of professional players and the ordinary people appears as a slight difference in the probability distribution,  it is a kind of evidence that the professional players have more ability to look far away. This subtle difference shows the difference in skills between professional players and the ordinary people who can play Go as a hobby. Professional players are those who have been practicing Go for a long time to overcome the subtle differences. However, AlphaGo has a high probability of placement of a long-distance, which is more clearly distinguished than professional players. It's a very simple and distinctive demonstration to show that AlphaGo's Go skills far exceeded human skills.

\begin{figure}
\onefigure[width=0.45\textwidth]{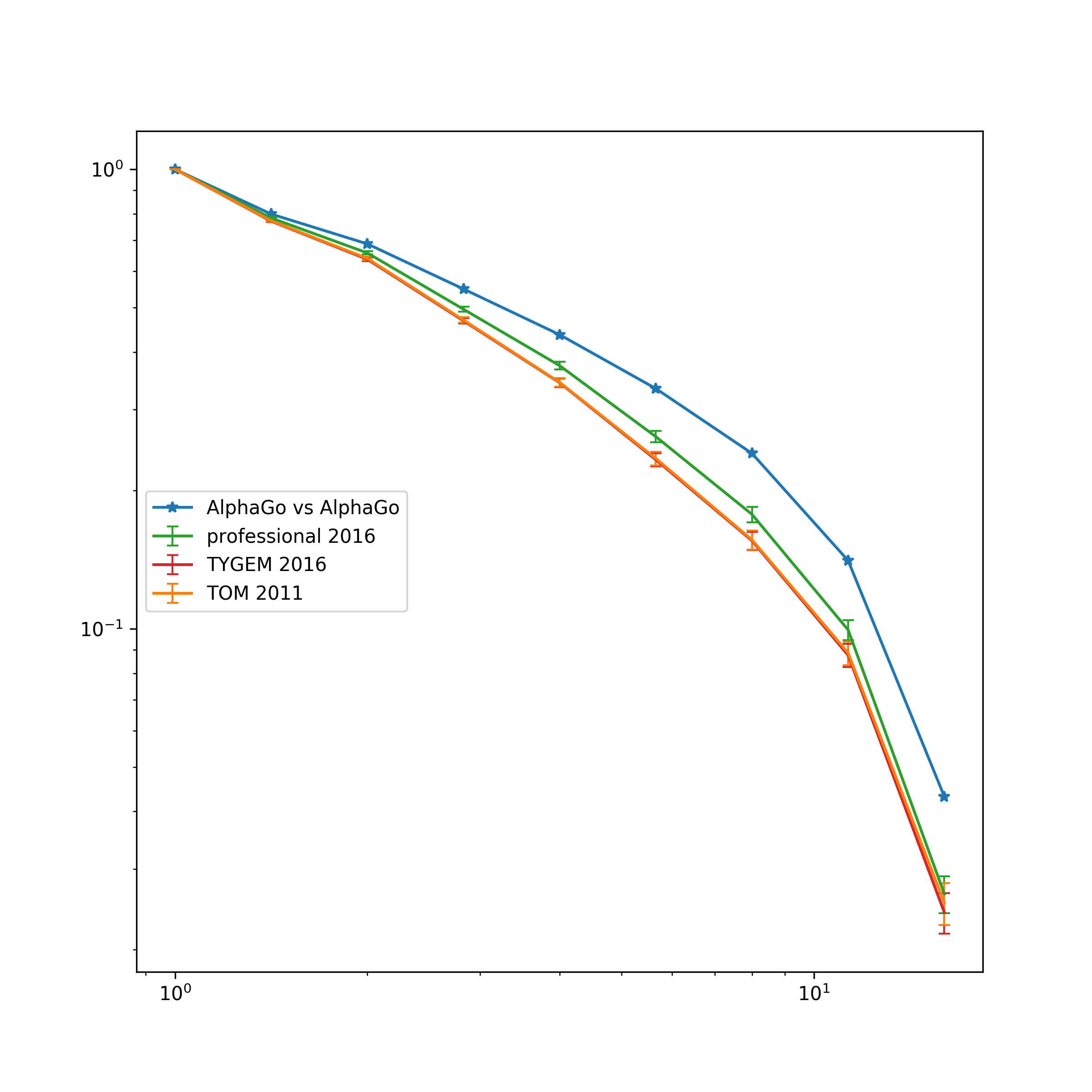}
\caption{Mean and variance for 10,000 iterations of the cumulative probability distributions for 55 samples among all game records in 2016 for professional matches, in 2016 for TYGEM, and in 2011 for TOM.}
\label{fig3}
\end{figure}

The number of games of TYGEM, TOM, and professional players is quite large, as shown in Tables \ref{table1} and \ref{table2}. As a result, there are occasional matches that show a significant long-range likelihood of being seen in AlphaGo's game, but the odds may have been lowered the likelihood by numerous other games. Thus, when we observed 55 games by chance in a game between humans, we checked that the same results like 55 self-games between AlphaGo could occur. The probability distribution was calculated from randomly selected 55 games from 167,085 games in TYGEM of 2016, 6,627 games in TOM of 2011, and 3,380 games in professional players of 2016.
This probability distribution was generated 10,000 times independently for each case. The mean of 10,000 probability distributions and the magnitude of their variance can be seen in Fig. \ref{fig3}. The probability distribution which is occurred from the games of professional players seems  unlikely to occur in the game of ordinary people. In addition, it is also extremely unlikely that the probability distributions of AlphaGo's games will occur in the games of the Professional player's games. In the probability distribution of consecutive placement distances, it is judged that the gap between the ordinary people and the professional players and the clear gap between the professional players and AlphaGo did not occur by chance.

\section{Conclusion}
One of Go's successful strategies is ``着眼大局 着手小局''. If you read it directly, it tells you that keep your eyes on the entire board and your hands on the board. It means that you should read the whole game and look far and at the same time, you should focus on the small part when you placing stone. This strategy inspired us to look at the balance between the long-distance and near-distance consecutive placement of stones. It is expected that while frequent short-distance action occurs focusing on the fight,  sometimes long-distance action occurs looking at the entire board. This behavior makes the probability distribution over the distance of successive placement forms the power-law function\cite{newman2005}. In this probability distribution analysis, we can see from the thicker tail that the ability of professional players to see the whole board and look far away is higher than the general people. Moreover, it is found a sign that AlphaGo has gone far exceed beyond the level of such professional players. This study shows that it is possible to gauge the level of the Go ability with very simple statistics without the need for a complicated evaluation by the Go expert.

If we have a sequential Go game record of AlphaGo from the beginning of learning to the completion of learning, it may be possible to index the growth of the skills by the method introduced in this study. And creating an abstract dynamic model like Ising model \cite{alvarado2017, rojas2019} that can reproduce Go's consecutive placement distance probability distribution would be interesting further work. The model could be used to verify that the ability to look far is an important factor for winning in Go game.

\acknowledgments
We would like to thank Dong Gun Kam very much for the introduction and advice about Go record data.

\end{document}